\documentclass{article}

\usepackage{arxiv}

\usepackage[utf8]{inputenc} 
\usepackage[T1]{fontenc}    
\usepackage{hyperref}       
\usepackage{url}            
\usepackage{booktabs}       
\usepackage{amsfonts}       
\usepackage{nicefrac}       
\usepackage{microtype}      
\usepackage{lipsum}
\usepackage{tabularx}
\usepackage{graphicx}
\graphicspath{ {./figs/} }

\title{Cross-Modal Learning of \\Housing Quality in Amsterdam}

\author{
 Alex Levering\\
  Wageningen University\\
  Wageningen, the Netherlands \\
  \texttt{ahlevering@gmail.com} \\
   \And
Diego Marcos \\
  INRIA\\
  Montpellier, France \\
  \And
 Devis Tuia \\
  EPFL\\
  Sion, Switzerland \\
}

\begin{document}
\maketitle

\begin{abstract}
In our research we test data and models for the recognition of housing quality in the city of Amsterdam from ground-level and aerial imagery. For ground-level images we compare Google StreetView (GSV) to Flickr images. Our results show that GSV predicts the most accurate building quality scores, approximately 30\% better than using only aerial images. However, we find that through careful filtering and by using the right pre-trained model, Flickr image features combined with aerial image features are able to halve the performance gap to GSV features from 30\% to 15\%. Our results indicate that there are viable alternatives to GSV for liveability factor prediction, which is encouraging as GSV images are more difficult to acquire and not always available. 
\end{abstract}

\section{Introduction}
Modern-day urbanization has led to large increases in the number of people living in cities. It is expected that more than half of the global population will live in cities by 2050 \cite{united_nations_department_of_economic_and_social_affairs_population_division_world_2019}. While cities are increasingly important for finding work, it is frequently the case that cities show very disparate access to service and quality of infrastructure, and therefore mixed quality of life in urban dwellings. Not taking into account people's social and physical housing needs can have detrimental effects on their well-being. For instance, the physical quality of housing is an indicator for one's mental well-being \cite{evans_built_2003}. In a broader sense, the quality of a neighbourhood may affect the residents' dietary and physical activity patterns \cite{thompson_healthy_2014}, as well as their morbidity \cite{barber_neighborhood_2016}. Evidently, monitoring that neighbourhoods are liveable and of an adequate quality could support policy makers and urban planners to design more liveable cities. Liveability is typically measured using surveys. However, surveyed data is expensive to acquire, infrequently available, and their results may be hard to scale beyond the original survey area. Ideally, quality of life data gathered through such surveys would be available at large scales, on a frequent basis and at low cost to monitor the liveability of urban areas and identify areas for improvement.

Image data such as ground-level photography can offer a solution to this problem, as it is easier to acquire and scale. Prior research has shown that ground-level images can reliably pick up attributes relating to urban sentiments \cite{dubey_deep_2016, naik_streetscore_2014}. A potential drawback is that large-scale collection of this data is often not trivial, and images may be affected by biases such as lighting and weather effects. Aerial images are another source of image data which can be considered. Their main advantage over ground-level images and surveys is that they can be used to survey large areas in a single data collection effort. For aerial images, it has also been proven that they can be used to survey factors relating to quality of life \cite{scepanovic_jane_2021, levering_liveability_2021}.

In this research we focus on building quality scores surveyed in Amsterdam at hectometer scale. We are interested in determining if a combination of ground-level images and aerial images can improve the prediction of liveability factors. For the aerial image modality, we train models using high-resolution aerial image data. For the ground-level model we compare two pre-trained feature extractors to determine if models tuned towards liveability make a noticeable difference in performance on our dataset of housing quality. Furthermore, we also train models to combine both modalities to test whether or not they can improve the overall prediction accuracy of housing quality.

\section{Data}
We use three data sources in our study: housing quality scores of the city of Amsterdam, Aerial imagery, and ground-level imagery.
\newline

\noindent\textbf{Housing Quality Scores}\newline
For our liveability ground truth labels we use housing quality scores over the city of Amsterdam. This score quantifies how housing contributes to liveability. This data is available as a grid with cells covering 100m$^{2}$ each. It is derived from various statistics such as building age, ownership situation, and consumption of utilities such as electricity. The statistics used to create the building score are averaged from buildings within 200m$^{2}$ from the patch center. The grid with scores is published by the Leefbaarometer project~\footnote{https://www.leefbaarometer.nl/}.\newline

\noindent\textbf{Aerial Imagery}\newline
For the aerial images we use aerial image patches of 500x500 pixels with a spatial resolution of 1 meter derived from the 2017 national aerial image dataset \cite{pdok_nieuw_2017}. For each patch, we consider the housing quality score attributable to the 100 meters patch center. We do this to ensure that the model has the context needed to recreate the housing score, as the scores were created by using data from a 200m$^{2}$ square meter radius from the cell center. As such, there is a 200m$^{2}$ overlap for each patch with its neighbouring patches.
\newline

\noindent\textbf{ground-level Images}\newline
We test two data types for our ground-level images. Firstly, we use Google StreetView (GSV) panorama images, which are widely used for urban attribute prediction \cite{naik_streetscore_2014, dubey_deep_2016}. We use the dataset of GSV panorama images in Amsterdam from \cite{srivastava_understanding_2019}. The dataset is designed for building function classification, and as such each panorama image in this dataset is oriented to directly face a building in the city by using the location and orientation data of the panorama images. After filtering images to the extents of the aerial image patches, we retained $90'256$ images. Our second dataset consists of Flickr images. Flickr is freely available and consists of crowdsourced images, making it more flexible and easier to acquire than GSV panorama images. It is a source of data that is increasingly used to study the environmental factors contributing to individuals' well-being from a first-person perspective \cite{havinga_defining_2020}. We gathered Flickr images taken between 2004 and 2020 with a geotag located in the city of Amsterdam, which resulted in 54'250 images.
\newline

\begin{figure}
    \centering
    \includegraphics[width=80mm,height=50mm]{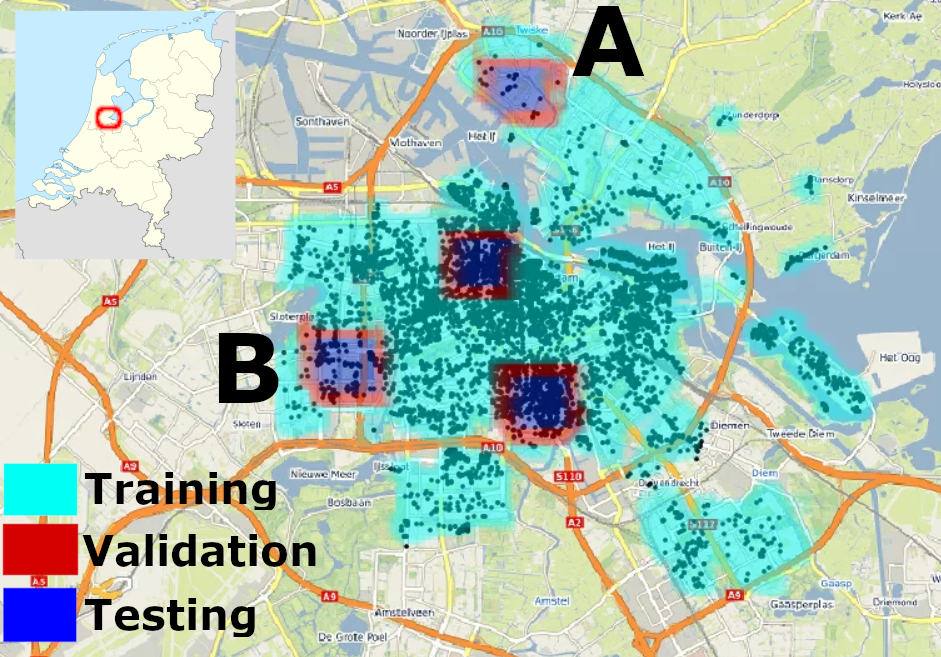}
    \caption{Data splits of our experiments over the city of Amsterdam. Testing squares are padded by validation cells to ensure that no test data is seen during training. Cyan squares are for training, red for validation, and blue for testing. Black points represent geotagged photos of the Flickr buildings subset.}
    \label{fig:splits}
\end{figure}

\noindent\textbf{Data splits}\newline
We split our dataset into training, validation and test sets by selecting square regions within the dataset. The edges of these squares partly overlap with the training set as a result of the patch size. We therefore assign the edges to the validation set. The region centers are assigned to the test set to avoid correlation between the sets due to spatial co-location. Our splits are shown in Figure~\ref{fig:splits}. For both sources of ground-level images, if no images intersect with an aerial image patch, then we leave the patch out of the subset.

\section{Methods}
\begin{figure}[!t]
    \centering
    \includegraphics[width=100mm,height=40mm]{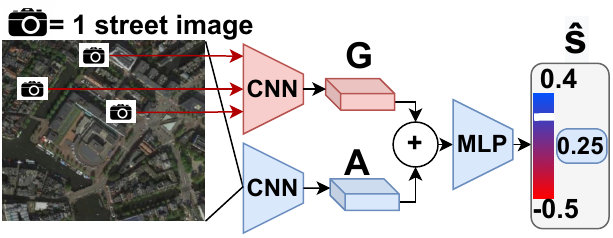}
    \caption{Multimodal model predicting housing quality scores. Only the aerial branch and the merging layer shown in blue are trained. Features extracted from the ground-level images are fixed. Depending on the subset, the ground-level image branch uses features extracted from either Google StreetView, or Flickr images.}
    \label{fig:method}
\end{figure}

Our model is tasked with predicting a housing quality score $\hat{s}$ from data within a patch. It consists of an aerial feature branch and a ground-level feature branch, as shown in Figure~\ref{fig:method}. The ResNet-50~\cite{he_deep_2016} feature extractor of the aerial feature branch is initialized with weights for housing quality prediction over The Netherlands from our preliminary study \cite{levering_liveability_2021}. For ground-level features, we use a ResNet-50 pre-trained ImageNet model, as well as a pre-trained ResNet-50 Place Pulse 2 (PP2) \cite{dubey_deep_2016} model. Place Pulse 2 is a dataset for urban sentiment analysis consisting of GSV images. The aerial feature branch extracts a 2048-dimensional vector $\mathbf{a}$ from an input aerial image. The ground-level feature branch produces one 2048-dimensional feature vector for each of the $N$ geotagged ground-level images within the patch. We average-pool these vectors to form the ground-level feature vector vector $\mathbf{g}$, following the same design as \cite{srivastava_fine-grained_2020}. To merge the feature vectors $\mathbf{a}$ and $\mathbf{g}$ into the feature vector $\mathbf{m}$, we perform pairwise addition:

\begin{equation}
    \label{eq:merge}
    \mathbf{m} = \mathbf{a} + \frac{1}{N}\sum_{n=1}^{N}\mathbf{g}_{n}
\end{equation}

The vector $\mathbf{m}$ is then passed to a two-layer perceptron to extract joint features over the merged vector. We first apply batch normalization to the features, which are then passed to the first fully-connected layer which produces a 100-dimensional vector. These features are subsequently passed to the final fully-connected layer to regress the building score $\hat{s}$. We train our model using a Mean Squared Error loss calculated over the predicted patch housing score $\hat{s}$ w.r.t. the ground truth patch housing score $s$: 

\begin{equation}
    \mathcal{L}_{score} = (s-\hat{s})^2
    \label{eq:l_score}
\end{equation}

During the first three epochs, only the fully connected layers are trained. Starting at epoch four, the aerial feature extractor is also optimized to fine-tune it to Amsterdam. The ground-level feature extractor is not modified at any stage, since it has been pre-trained with images of a similar nature.

\section{Experimental setup}
Beyond reporting the results on the full method using either GSV or one of the Flickr subsets for the ground-level branch, we also perform ablation studies to test the performance of each branch individually. To test the aerial branch, we set the ground-level features $\mathbf{g}$ to be a vector containing zeroes. We do the same to the aerial features $\mathbf{a}$ to test the ground-level features.

We train all models with the Adam optimizer for 25 epochs, which we initialize with a learning rate of 0.001 and a weight decay of 0.001. We report the root mean squared error (RMSE) of the housing quality score, as well Kendall's $\tau$ \cite{kendall_new_1938}, which is a ranking coefficient between -1 and 1 which indicates whether or not samples are correctly placed in the right order in terms of increasing housing quality score. A value of -1 indicates a perfectly inverse ranking, while a value of 1 represents a perfect ranking.\newline 

\noindent\textbf{Filtering Flickr images }\newline
As Flickr consists of social media photos that are less organised than GSV images, filtering is necessary to retain only images which are beneficial for building score regression. We apply a pre-trained Places365 \cite{zhou_places_2018} model for scene classification to test two filtering methods. As a weak filtering method, we retain only images where 9 out of the 10 most activated classes are marked as outdoors in the dataset. We refer to this subset as \textbf{Flickr Outdoors}. In total, the Flickr Outdoors subset contains 34'222 images. Secondly, we select images that have at least one scene strongly related to buildings above a given threshold, for which we consider 24 building-related classes. This threshold was empirically tested and set to an activation of 0.05. This will filter the dataset more aggressively to focus on buildings in favour of the housing quality score. We refer to this subset as \textbf{Flickr Buildings}. The Flickr Buildings subset contains 11'774 images. Filtering out images also resulted in some patches having no ground-based images, which had to be excluded. We show our patch distribution per subset in Table~\ref{tab:n_samples}.

\begin{figure*}[t]
    \begin{center}
        \begin{tabular}{cccccc}
            \includegraphics[width=33mm,height=22mm]{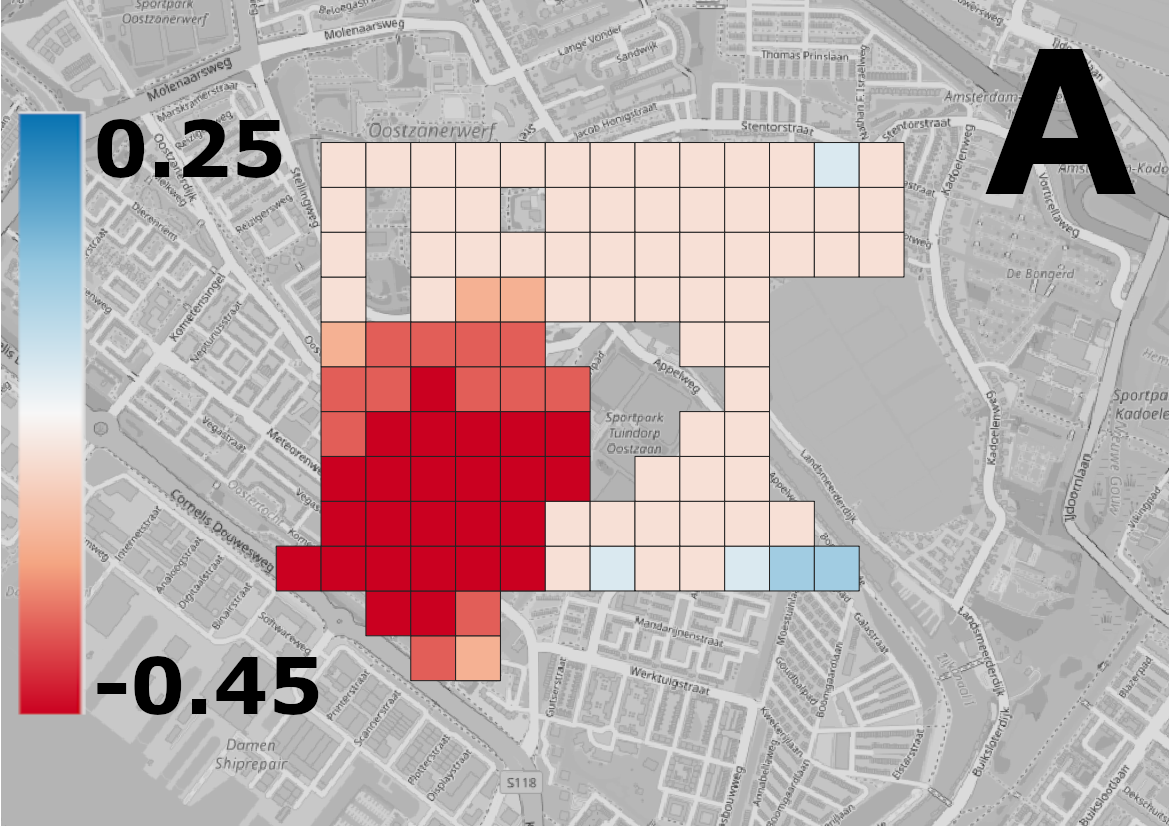}&     
            \includegraphics[width=33mm,height=22mm]{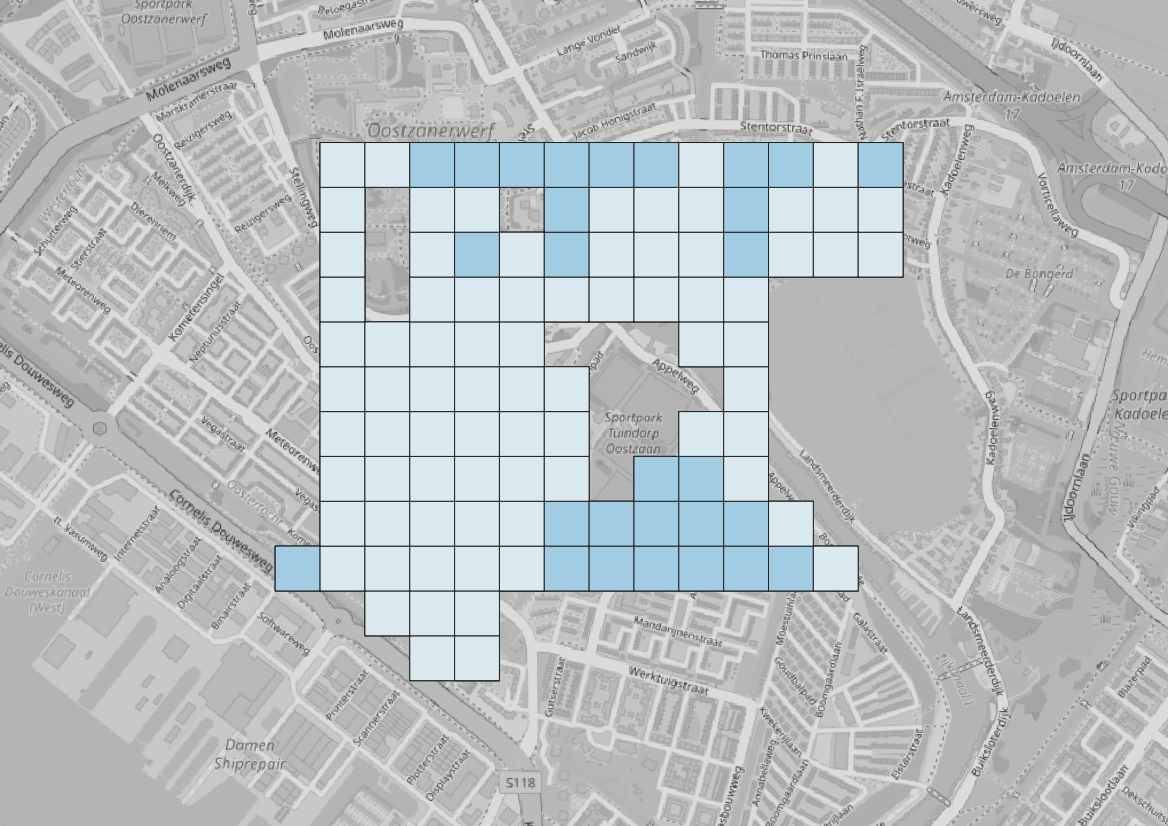}&
            \includegraphics[width=33mm,height=22mm]{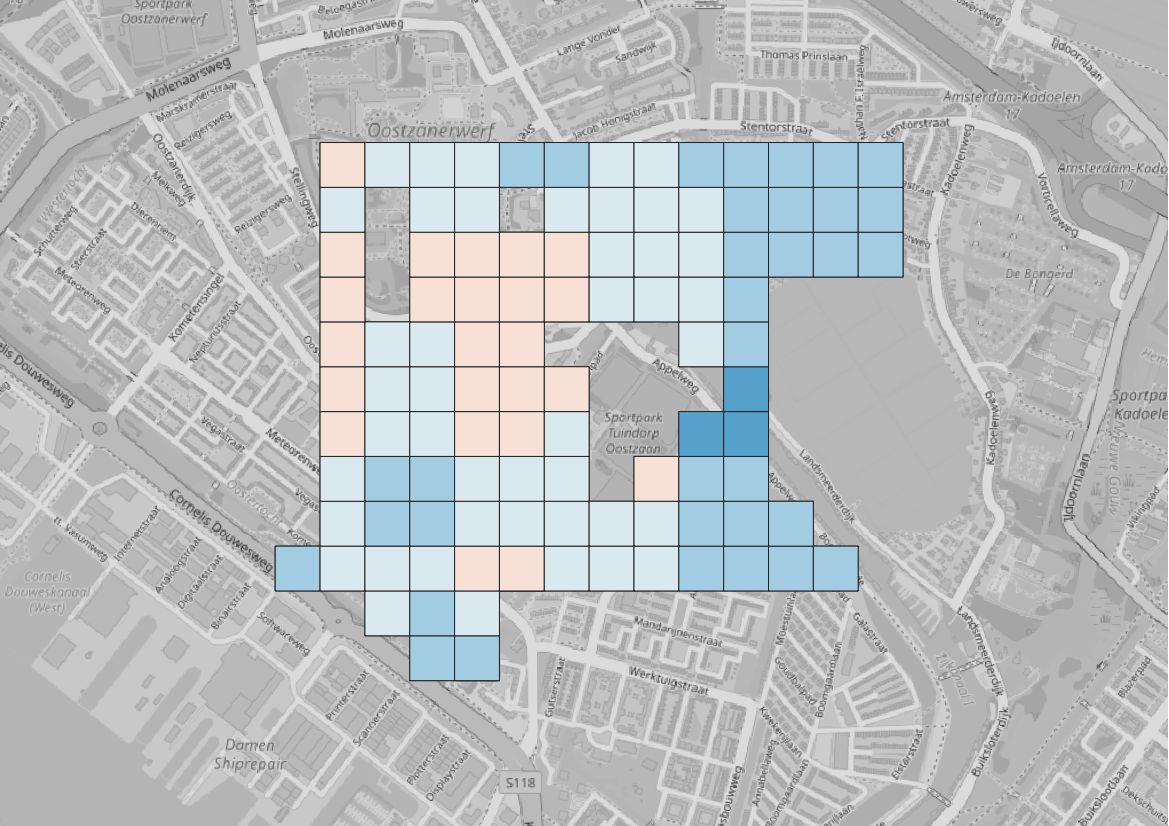}&
            \includegraphics[width=33mm,height=22mm]{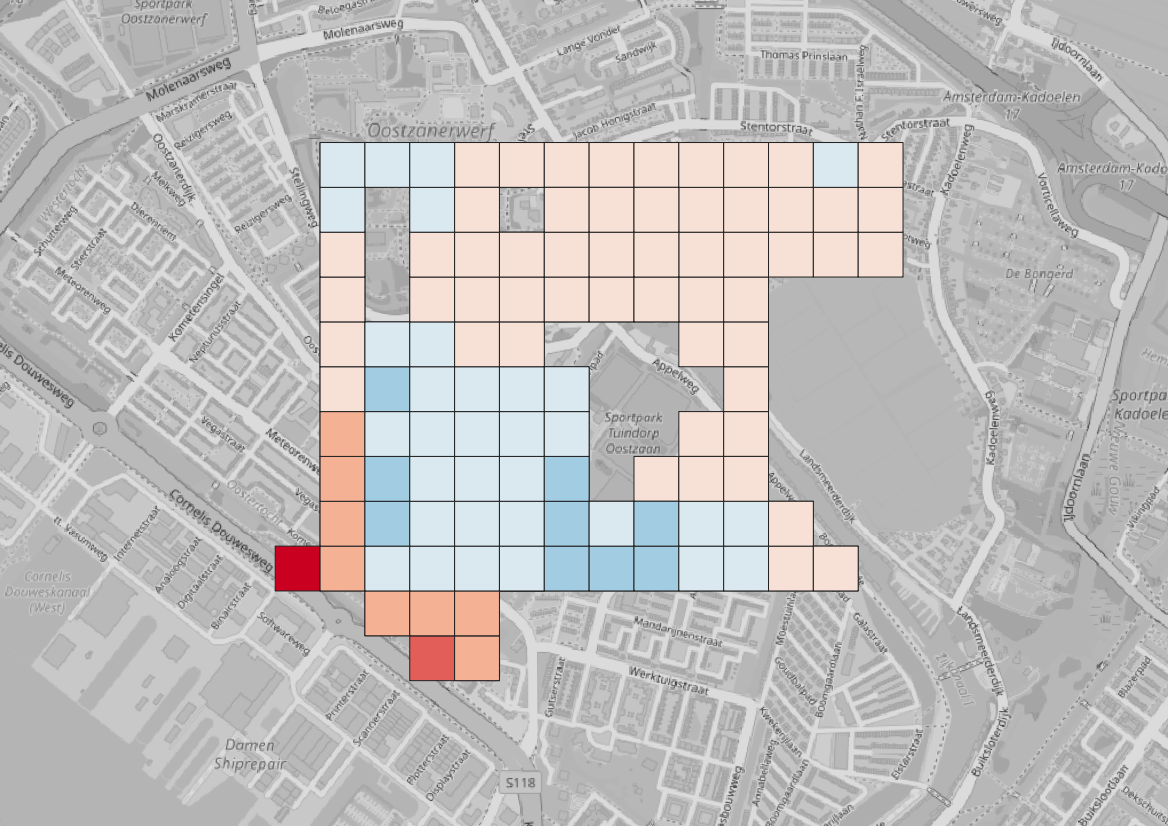}&
            \includegraphics[width=33mm,height=22mm]{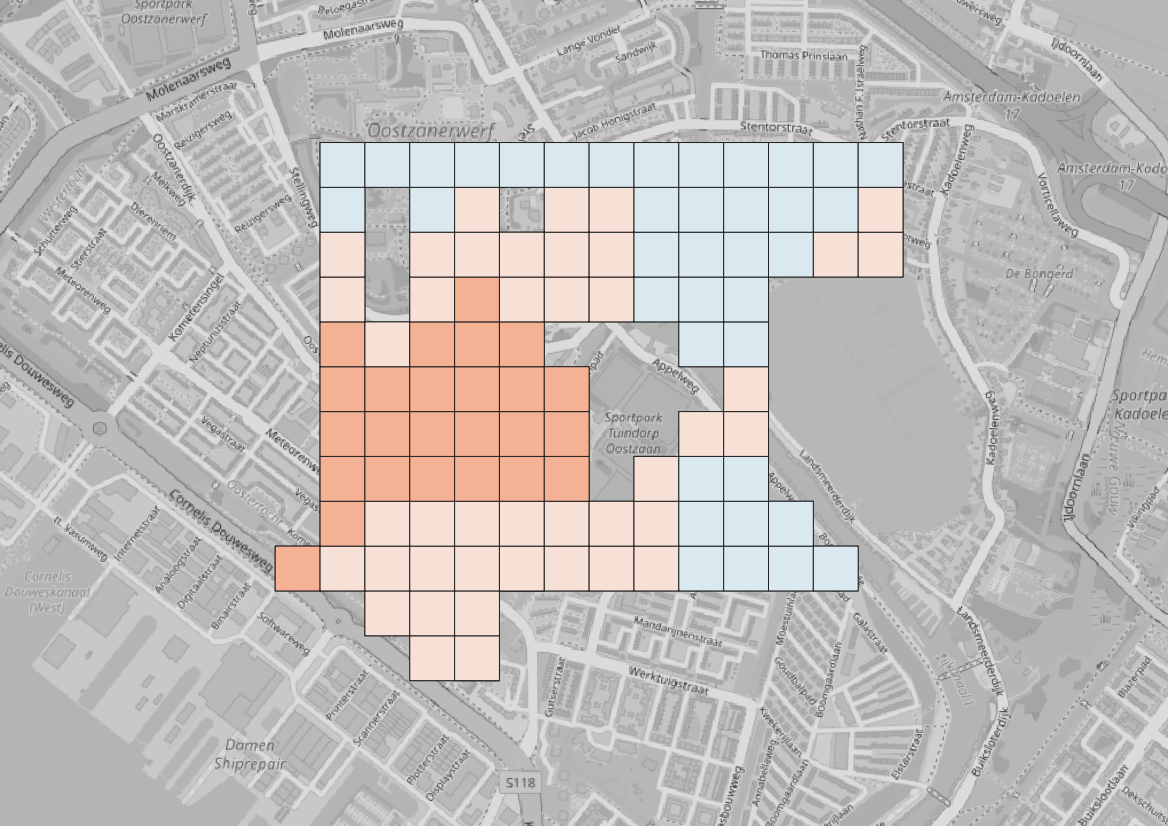}\\
            Ground Truth & Only Aerial & Outdoors PP2 + Aerial & Buildings PP2 + Aerial & GSV Only PP2 \\
        \end{tabular}
        \begin{tabular}{cccccc}
            \includegraphics[width=33mm,height=22mm]{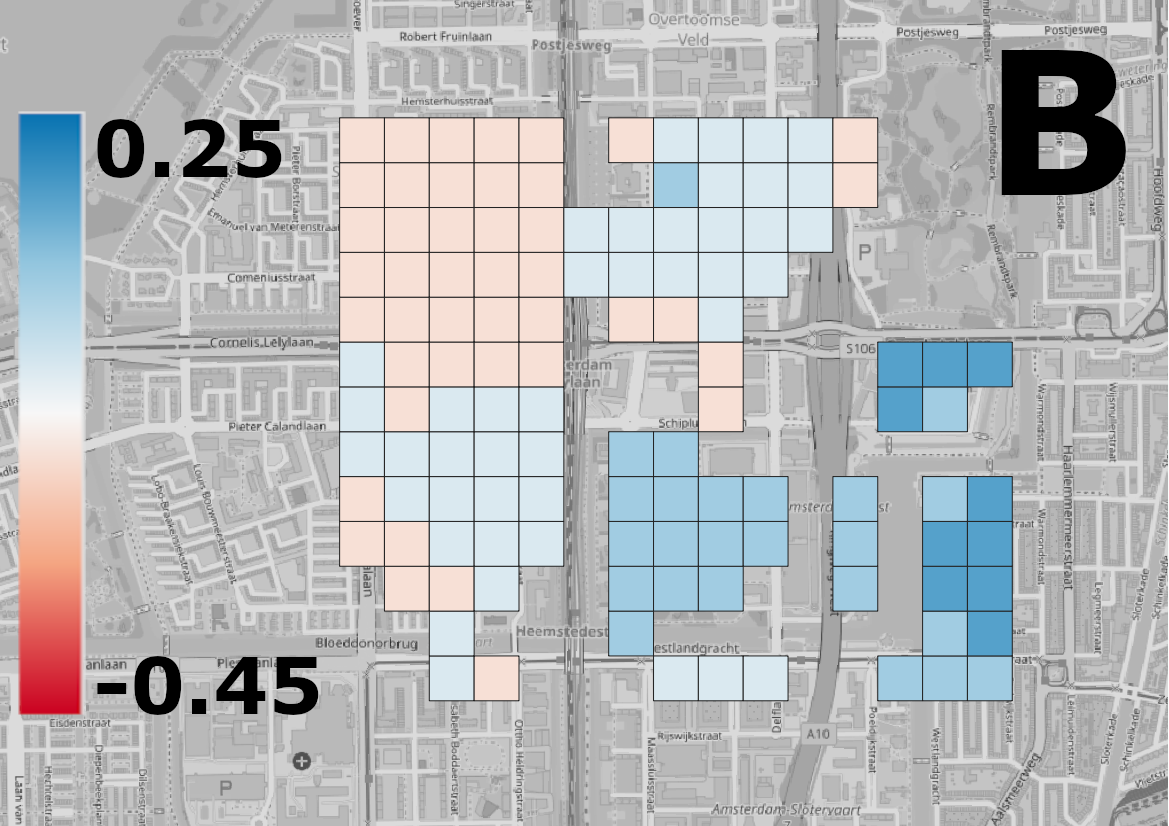}&     
            \includegraphics[width=33mm,height=22mm]{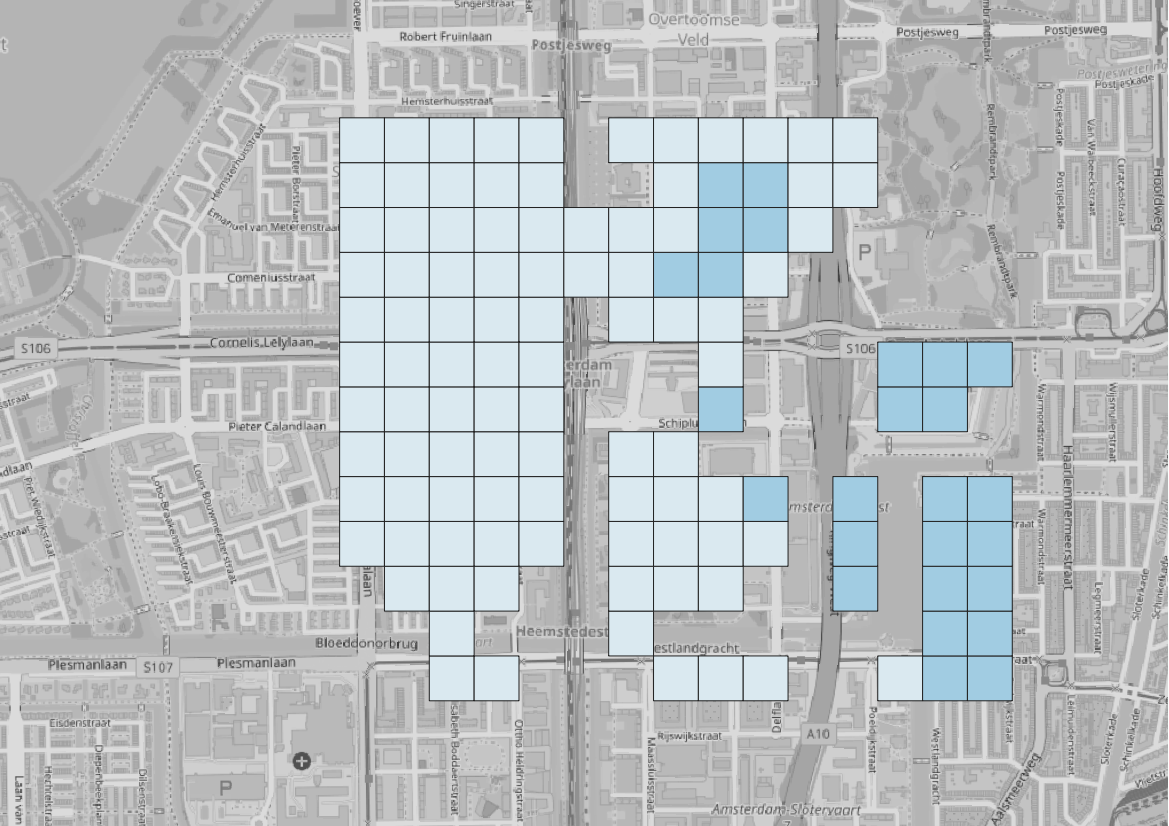}&
            \includegraphics[width=33mm,height=22mm]{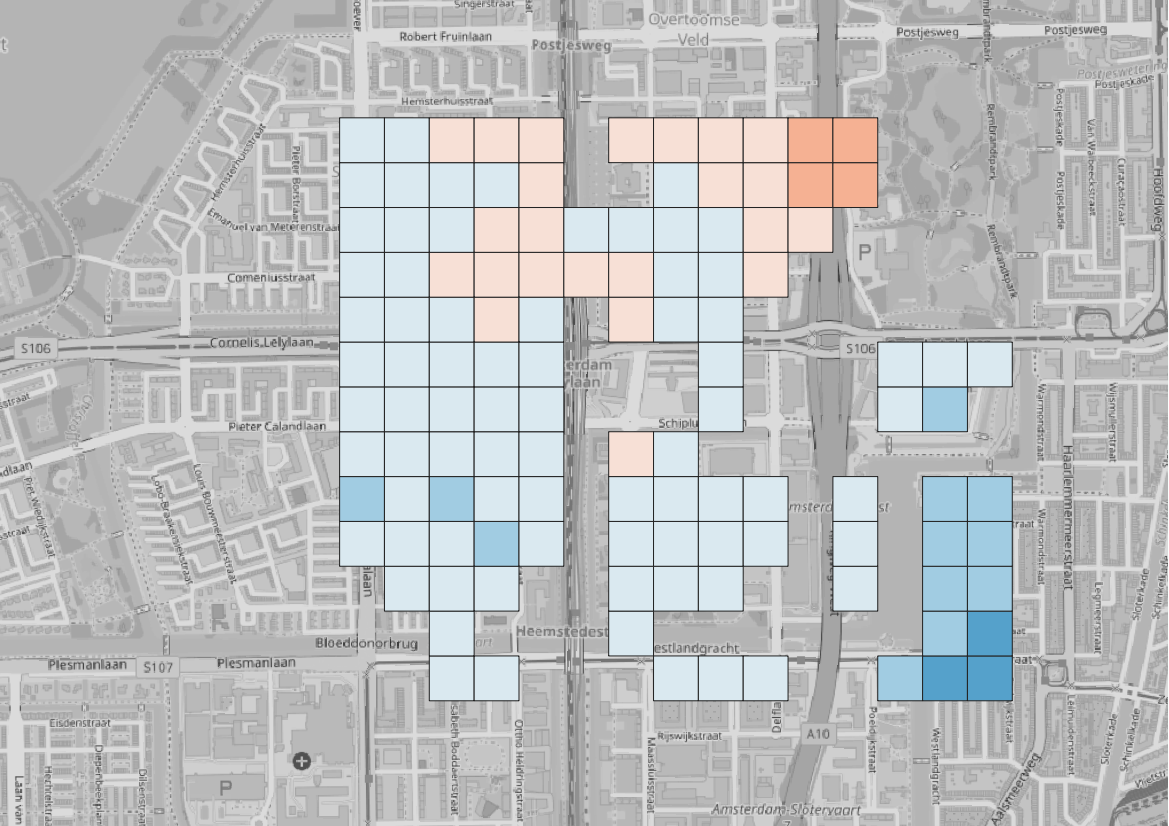}&
            \includegraphics[width=33mm,height=22mm]{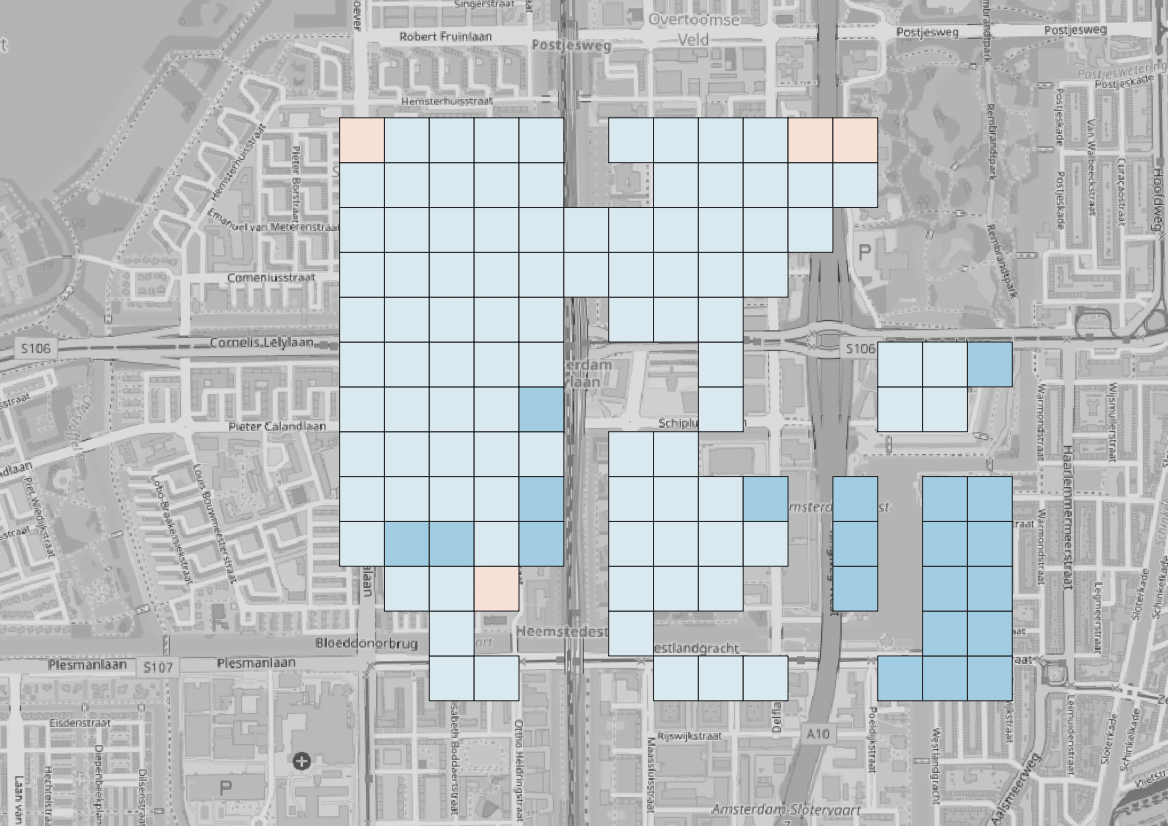}&
            \includegraphics[width=33mm,height=22mm]{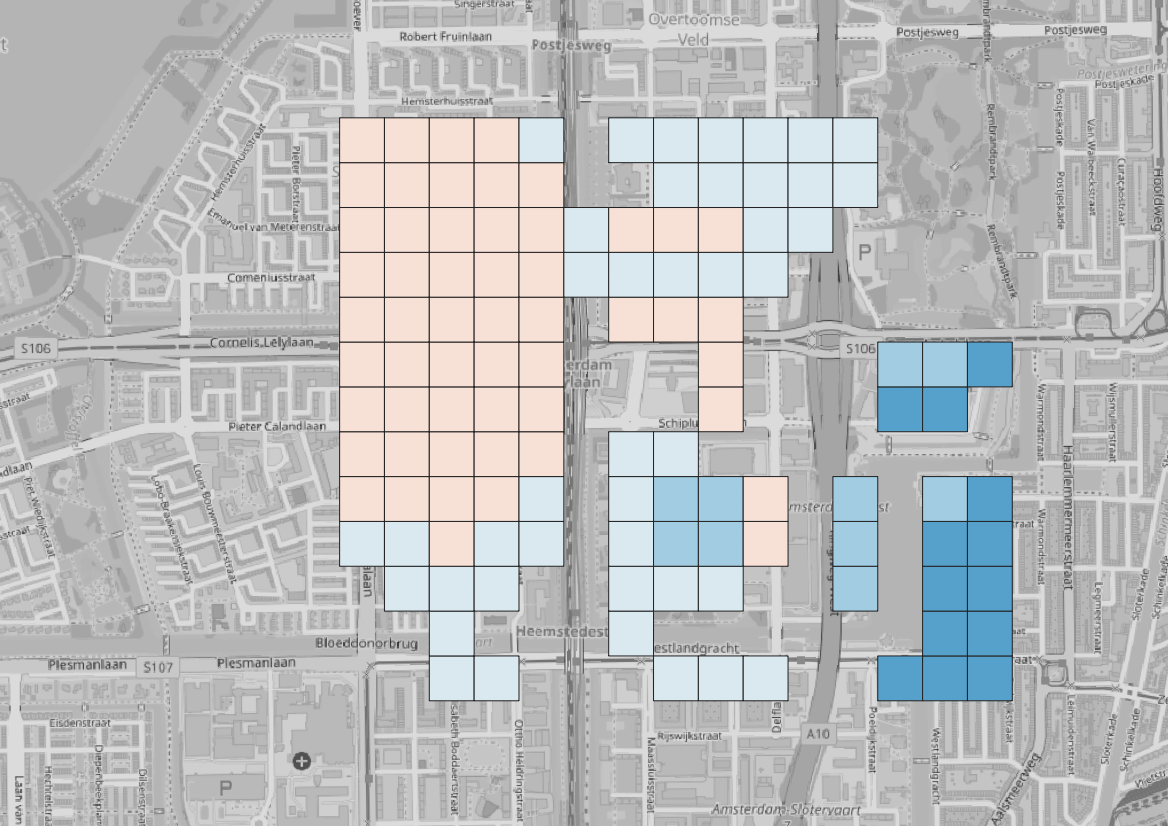}\\
            Ground Truth & Only Aerial & Outdoors PP2 + Aerial & Buildings PP2 + Aerial & GSV Only PP2 \\
        \end{tabular}     
    \end{center}
    \caption{Plots of predictions of building quality score for the best model of each data subset on the two most spatially diverse tiles. Their locations are displayed in Figure~\ref{fig:splits}. Colors range from red (low-quality) to blue (high-quality).}
    \label{fig:maps}
\end{figure*}

\section{Results and Discussion}
\noindent\textbf{Using GSV images as ground modality }\newline
 We show our results when using GSV images for the ground branch in Table~\ref{tab:gsvs}. Overall, we find that using only features extracted using a PP2 model provides the best results, slightly edging out ImageNet features in terms of Kendall's $\tau$, which reaches a value of $0.778$. When using GSV for ground-level features, merging with overhead aerial imagery does not provide performance improvements regardless of the feature extractor. In addition, pre-training the feature extractor on ImageNet or PP2 results in similar performances. Compared to a Kendall's $\tau$ of $0.5810$ obtained from using only aerial images, there is a 30\% performance gap.
\vspace{20pt}

\begin{table}
  \caption{Patches per split for each subset}
  \centering
  \label{tab:n_samples}
  \begin{tabularx}{0.525\columnwidth}{lcccc}
    \toprule
    \textbf{Subset} & \textbf{Train} & \textbf{Validation} & \textbf{Test} & \textbf{Coverage}\\
    \midrule
    Aerial              & 4'300 & 570 & 491 & 100\%     \\    
    GSV                 & 4'294 & 570 & 491 & 99.88\%   \\
    Flickr Outdoors     & 4'255 & 570 & 491 & 98.97\%   \\
    Flickr Buildings    & 4'027 & 538 & 458 & 93.69\%   \\
  \bottomrule
\end{tabularx}
\end{table}

\noindent\textbf{Using Flickr images as ground modality}\newline
Table~\ref{tab:flickr} shows the results obtained from the Flickr ground images. Our results show substantial differences between the three modalities. The least competitive result in terms of Kendall's $\tau$ occurs when merging the aerial image features with Outdoor Flickr images using ImageNet features, $0.576$. The best unimodal setting with Flickr images consists of using PP2 features from the Flickr Buildings subset, reaching a Kendall's $\tau$ of $0.649$. For unimodal prediction it is important to simultaneously use the Flickr Buildings images along with PP2 pre-training, since using either the Flickr Outdoors subset or only ImageNet pre-training results in a loss of performance, down to $0.596$ and $0.602$ respectively. Adding the aerial branch to the PP2 pre-trained Flickr Buildings model result in another increase in performance, up to $0.686$. By using a combination of Flickr Buildings PP2 features and the aerial features, the performance gap compared to the best GSV model is halved.

By comparing the metrics of the three subsets, we can assess the suitability of alternatives to be used instead of GSV. While GSV image features prove to be most suitable for building quality at the city scale, it is encouraging that Flickr and aerial images are able to close the performance gap. Furthermore, while GSV images are more suitable for urban analyses, the data is often difficult to acquire for larger areas, or even entirely unavailable. In contrast, Flickr images are easy to acquire and widely available. The success of using Flickr shows that general-purpose social media data sources can also be used, as they are easier to scale over larger areas, for instance through crowdsourcing efforts.\newline

\noindent\textbf{Spatial Predictions}\newline
In Figure~\ref{fig:maps} we show the spatial predictions for the best model of each subset for the two most spatially diverse testing tiles. The maps show that the PP2-only GSV model is able to approximate the ground truth most accurately in both tiles. Both Flickr models struggle with predicting the extent of the low quality housing of tile A. This may be caused by a lack of images in the north of Amsterdam, which can be seen in the black points of Figure~\ref{fig:splits}, which represent images of the Buildings subset. This area is less popular with tourists, which may explain the lack of Flickr photos. The GSV dataset has much better coverage in this area, which is reflected by the better prediction quality for this tile.

\begin{table}
  \caption{Metrics when using GSV as ground image source}
  \label{tab:gsvs}
  \centering
  \begin{tabularx}{0.525\columnwidth}{lXcc}
    \toprule
    \textbf{Modality} & \textbf{Ground Features} & \textbf{RMSE} & \textbf{Kendall's $\tau$}\\
    \midrule
    Aerial          & n.a. & 0.1314 & 0.5818 \\    
    GSV             & ImageNet & \textbf{0.0670} & 0.7651\\
    GSV             & PP2 & 0.0707 & \textbf{0.7780} \\
    Aerial \& GSV    & ImageNet & 0.0787 & 0.7699 \\
    Aerial \& GSV    & PP2 & 0.0765 & 0.7656\\
  \bottomrule
\end{tabularx}
\end{table}

\begin{table}
  \caption{Metrics when using Flickr as ground image source}
  \centering
  \label{tab:flickr}
  \begin{tabularx}{0.575\columnwidth}{lccc}
    \toprule
    \textbf{Modality} & \textbf{Ground Features} & \textbf{RMSE} & \textbf{Kendall's $\tau$}\\
    \midrule
    Aerial              & n.a       & 0.1314 & 0.5810 \\
    Outdoor             & ImageNet  & 0.1100 & 0.5755 \\
    Outdoors            & PP2       & 0.1155 & 0.5962 \\
    Buildings           & ImageNet  & 0.1179 & 0.6024 \\
    Buildings           & PP2       & \textbf{0.1031} & 0.6487 \\
    Aerial \& Outdoors  & ImageNet  & 0.1427 & 0.5729 \\
    Aerial \& Outdoors  & PP2       & 0.1306 & 0.6215 \\
    Aerial \& Buildings & ImageNet  & 0.1142 & 0.6243 \\
    Aerial \& Buildings & PP2       & 0.1104 & \textbf{0.6862} \\
  \bottomrule
\end{tabularx}
\end{table}

\section{Conclusions}
In this paper we use a combination of features extracted from ground-level and aerial images to predict the quality of houses in Amsterdam. For the ground-level images we tested two pre-trained feature extractors, one pre-trained on ImageNet, and one on Place Pulse 2, a dataset for subjective perception of urban ground-level images. We collected and refined three ground-level image datasets: Google Streetview (GSV), Flickr Outdoors and Flickr Buildings. The latter two were obtained by filtering geotagged Flickr images. Using only GSV images resulted in the best overall performance, providing a $30\%$ increase in Kendall's $\tau$ with respect to using only aerial imagery. This suggests that the nature of GSV imagery is well-suited, as it captures $360^\circ$ panoramas of most of the city's streets at regular intervals. However, this type of imagery is costly to obtain and not always available. Our results show that using less curated but more easily available social media images such as Flickr can still provide a $15\%$ increase in performance w.r.t. the aerial imagery if both the images and the feature extractor are carefully selected for the task.



\begin{thebibliography}{10}
\providecommand{\url}[1]{#1}
\csname url@samestyle\endcsname
\providecommand{\newblock}{\relax}
\providecommand{\bibinfo}[2]{#2}
\providecommand{\BIBentrySTDinterwordspacing}{\spaceskip=0pt\relax}
\providecommand{\BIBentryALTinterwordstretchfactor}{4}
\providecommand{\BIBentryALTinterwordspacing}{\spaceskip=\fontdimen2\font plus
\BIBentryALTinterwordstretchfactor\fontdimen3\font minus
  \fontdimen4\font\relax}
\providecommand{\BIBforeignlanguage}[2]{{%
\expandafter\ifx\csname l@#1\endcsname\relax
\typeout{** WARNING: IEEEtran.bst: No hyphenation pattern has been}%
\typeout{** loaded for the language `#1'. Using the pattern for}%
\typeout{** the default language instead.}%
\else
\language=\csname l@#1\endcsname
\fi
#2}}
\providecommand{\BIBdecl}{\relax}
\BIBdecl

\bibitem{united_nations_department_of_economic_and_social_affairs_population_division_world_2019}
{United Nations, Department of Economic and Social Affairs, Population
  Division}, ``World {Urbanization} {Prospects}: {The} 2018 {Revision},'' New
  York, Tech. Rep. ST/ESA/SER.A/420, 2019.

\bibitem{evans_built_2003}
G.~W. Evans, ``\BIBforeignlanguage{eng}{The built environment and mental
  health},'' \emph{\BIBforeignlanguage{eng}{J Urban Health}}, vol.~80, no.~4,
  pp. 536--555, Dec. 2003.

\bibitem{thompson_healthy_2014}
S.~Thompson and J.~Kent, ``Healthy {Built} {Environments} {Supporting}
  {Everyday} {Occupations}: {Current} {Thinking} in {Urban} {Planning},''
  \emph{Journal of Occupational Science}, vol.~21, no.~1, pp. 25--41, Jan.
  2014.

\bibitem{barber_neighborhood_2016}
S.~Barber, D.~A. Hickson, I.~Kawachi, S.~V. Subramanian, and F.~Earls,
  ``\BIBforeignlanguage{eng}{Neighborhood {Disadvantage} and {Cumulative}
  {Biological} {Risk} {Among} a {Socioeconomically} {Diverse} {Sample} of
  {African} {American} {Adults}: {An} {Examination} in the {Jackson} {Heart}
  {Study}},'' \emph{\BIBforeignlanguage{eng}{J Racial Ethn Health
  Disparities}}, vol.~3, no.~3, pp. 444--456, 2016.

\bibitem{dubey_deep_2016}
A.~Dubey, N.~Naik, D.~Parikh, R.~Raskar, and C.~A. Hidalgo,
  ``\BIBforeignlanguage{en}{Deep {Learning} the {City}: {Quantifying} {Urban}
  {Perception} at a {Global} {Scale}},'' in
  \emph{\BIBforeignlanguage{en}{{ECCV} 2016}}, B.~Leibe, J.~Matas, N.~Sebe, and
  M.~Welling, Eds.\hskip 1em plus 0.5em minus 0.4em\relax Cham: Springer, 2016,
  pp. 196--212.

\bibitem{naik_streetscore_2014}
\BIBentryALTinterwordspacing
N.~Naik, J.~Philipoom, R.~Raskar, and C.~Hidalgo,
  ``\BIBforeignlanguage{en}{Streetscore -- {Predicting} the {Perceived}
  {Safety} of {One} {Million} {Streetscapes}},'' in
  \emph{\BIBforeignlanguage{en}{{CVPR} 2014}}.\hskip 1em plus 0.5em minus
  0.4em\relax Columbus, OH, USA: IEEE, Jun. 2014, pp. 793--799. [Online].
  Available:
  \url{http://ieeexplore.ieee.org/lpdocs/epic03/wrapper.htm?arnumber=6910072}
\BIBentrySTDinterwordspacing

\bibitem{scepanovic_jane_2021}
S.~Scepanovic, S.~Joglekar, S.~Law, and D.~Quercia, ``Jane {Jacobs} in the
  {Sky}: {Predicting} {Urban} {Vitality} with {Open} {Satellite} {Data},''
  \emph{ACM Human-Computer Interaction}, vol.~5, pp. 1--25, Apr. 2021.

\bibitem{levering_liveability_2021}
A.~Levering, D.~Marcos, and D.~Tuia, ``Liveability from {Above}:
  {Understanding} {Quality} of {Life} with {Overhead} {Imagery} and {Deep}
  {Neural} {Networks},'' in \emph{Proceedings of {IGARSS} 2021}.\hskip 1em plus
  0.5em minus 0.4em\relax Brussels: IEEE, Jul. 2021.

\bibitem{pdok_nieuw_2017}
\BIBentryALTinterwordspacing
{PDOK}, ``{NIEUW}: hogere resolutie luchtfoto als open data bij {PDOK},'' Oct.
  2017. [Online]. Available:
  \url{https://www.pdok.nl/-/nieuw-hogere-resolutie-luchtfoto-als-open-data-bij-pdok}
\BIBentrySTDinterwordspacing

\bibitem{srivastava_understanding_2019}
\BIBentryALTinterwordspacing
S.~Srivastava, J.~E. Vargas-Muñoz, and D.~Tuia, ``Understanding urban landuse
  from the above and ground perspectives: a deep learning, multimodal
  solution,'' \emph{Remote Sensing of Environment}, vol. 228, pp. 129--143,
  Jul. 2019, arXiv: 1905.01752. [Online]. Available:
  \url{http://arxiv.org/abs/1905.01752}
\BIBentrySTDinterwordspacing

\bibitem{havinga_defining_2020}
\BIBentryALTinterwordspacing
I.~Havinga, P.~W. Bogaart, L.~Hein, and D.~Tuia,
  ``\BIBforeignlanguage{en}{Defining and spatially modelling cultural ecosystem
  services using crowdsourced data},'' \emph{\BIBforeignlanguage{en}{Ecosystem
  Services}}, vol.~43, p. 101091, Jun. 2020. [Online]. Available:
  \url{https://www.sciencedirect.com/science/article/pii/S2212041620300334}
\BIBentrySTDinterwordspacing

\bibitem{he_deep_2016}
K.~He, X.~Zhang, S.~Ren, and J.~Sun, ``Deep {Residual} {Learning} for {Image}
  {Recognition},'' in \emph{{CVPR 2016}}, Jun. 2016, pp. 770--778.

\bibitem{srivastava_fine-grained_2020}
\BIBentryALTinterwordspacing
S.~Srivastava, J.~E. Vargas~Muñoz, S.~Lobry, and D.~Tuia, ``Fine-grained
  landuse characterization using ground-based pictures: a deep learning
  solution based on globally available data,'' \emph{IJGIS}, vol.~34, no.~6,
  pp. 1117--1136, Jun. 2020. [Online]. Available:
  \url{https://doi.org/10.1080/13658816.2018.1542698}
\BIBentrySTDinterwordspacing

\bibitem{kendall_new_1938}
\BIBentryALTinterwordspacing
M.~G. Kendall, ``A {New} {Measure} for {Rank} {Correlation},''
  \emph{Biometrika}, vol.~30, no. 1-2, pp. 81--93, Jun. 1938. [Online].
  Available: \url{https://doi.org/10.1093/biomet/30.1-2.81}
\BIBentrySTDinterwordspacing

\bibitem{zhou_places_2018}
B.~Zhou, A.~Lapedriza, A.~Khosla, A.~Oliva, and A.~Torralba, ``Places: {A} 10
  {Million} {Image} {Database} for {Scene} {Recognition},'' \emph{IEEE TPAM},
  vol.~40, no.~6, pp. 1452--1464, Jun. 2018.

\end{thebibliography}
\end{document}